\documentclass[10pt,twocolumn]{article}

\usepackage{cvpr}
\usepackage{times}
\usepackage{epsfig}
\usepackage{graphicx}
\usepackage{amsmath}
\usepackage{amssymb}
\usepackage{xspace}
\usepackage{subcaption}
\usepackage{makecell}
\usepackage{xcolor}

\usepackage[pagebackref=true,breaklinks=true,colorlinks,bookmarks=false]{hyperref}

\cvprfinalcopy 

\def\cvprPaperID{3} 

\ifcvprfinal\fi

\newcommand{\mysection}[1]{\vspace{2pt}\noindent\textbf{#1}}

\newcommand\teacher{\ensuremath{\mathcal{T}}\xspace}
\newcommand\student{\ensuremath{\mathcal{S}}\xspace}
\newcommand\camteacher{\ensuremath{\mathcal{C}_\mathcal{T}}\xspace}
\newcommand\camstudent{\ensuremath{\mathcal{C}_\mathcal{S}}\xspace}

\newcommand\intersectFOV{\ensuremath{\mathbb{P}}\xspace}

\newcommand\notintersectFOV{\ensuremath{\overline{\intersectFOV}}\xspace}

\newcommand\capturetime{\ensuremath{t}\xspace}
\newcommand\frameteacher{\ensuremath{\camteacher(\capturetime)}\xspace}
\newcommand\framestudent{\ensuremath{\camstudent(\capturetime)}\xspace}

\newcommand\coordstudent{\ensuremath{(x,y,w,h)}\xspace}
\newcommand\radiusinitial{\ensuremath{r}\xspace}
\newcommand\radiusfinal{\ensuremath{r'}\xspace}
\newcommand\angleinitial{\ensuremath{\theta}\xspace}
\newcommand\anglefinal{\ensuremath{\theta'}\xspace}
\newcommand\polarinitial{\ensuremath{(\radiusinitial,\angleinitial)}\xspace}
\newcommand\polarfinal{\ensuremath{(\radiusfinal,\anglefinal)}\xspace}
\newcommand{\me}{\mathrm{e}}

\newcommand\maskVibe{\ensuremath{\text{M}(t)}\xspace}
\newcommand\bigmaskVibe{\ensuremath{\mathbb{M}(t)}\xspace}
\newcommand\notmaskVibe{\ensuremath{\overline{\text{M}(t)}}\xspace}
\newcommand\notbigmaskVibe{\ensuremath{\overline{\mathbb{M}(t)}}\xspace}

\newcommand\objectness{\ensuremath{p}\xspace}
\newcommand\tiou{\ensuremath{\text{t\_IoU}}\xspace}

\definecolor{anthoblue}[a=.5]{RGB}{36,130,181}
\definecolor{anthoorange}[a=.5]{RGB}{242,133,8}
\definecolor{anthogray}[a=.5]{RGB}{127,127,127}
\definecolor{anthogreen}[a=.5]{RGB}{82,140,28}
\definecolor{anthored}[a=.5]{RGB}{201,74,38}

\begin{document}

\title{Multimodal and multiview distillation \\ for real-time player detection on a football field}

\author{
Anthony Cioppa*\\
{\small University of Li\`ege}\\
{\tt\small anthony.cioppa@uliege.be}
\and
Adrien Deli\`ege*\\
{\small University of Li\`ege}\\
{\tt\small adrien.deliege@uliege.be}
\and
Noor Ul Huda\\
{\small Aalborg University}\\
{\tt\small nuh@create.aau.dk}
\and
Rikke Gade\\
{\small Aalborg University}\\
\phantom{{\tt\small adrien.deliege@uliege.be}}
\and
Marc Van Droogenbroeck\\
{\small University of Li\`ege}\\
\and
Thomas B. Moeslund\\
{\small Aalborg University}\\
}

\maketitle

\newcommand\blfootnote[1]{%
  \begingroup
  \renewcommand\thefootnote{}\footnote{#1}%
  \addtocounter{footnote}{-1}%
  \endgroup
}
\blfootnote{\textbf{(*)} Denotes equal contributions. Code at \url{https://github.com/cioppaanthony/multimodal-multiview-distillation}.}

\begin{abstract}
Monitoring the occupancy of public sports facilities is essential to assess their use and to motivate their construction in new places. In the case of a football field, the area to cover is large, thus several regular cameras should be used, which makes the setup expensive and complex. As an alternative, we developed a system that detects players from a unique cheap and wide-angle fisheye camera assisted by a single narrow-angle thermal camera. In this work, we train a network in a knowledge distillation approach in which the student and the teacher have different modalities and a different view of the same scene. In particular, we design a custom data augmentation combined with a motion detection algorithm to handle the training in the region of the fisheye camera not covered by the thermal one. We show that our solution is effective in detecting players on the whole field filmed by the fisheye camera. We evaluate it quantitatively and qualitatively in the case of an online distillation, where the student detects players in real time while being continuously adapted to the latest video conditions.
\end{abstract}


\begin{figure}[t]
    \centering
    \includegraphics[width=\columnwidth]{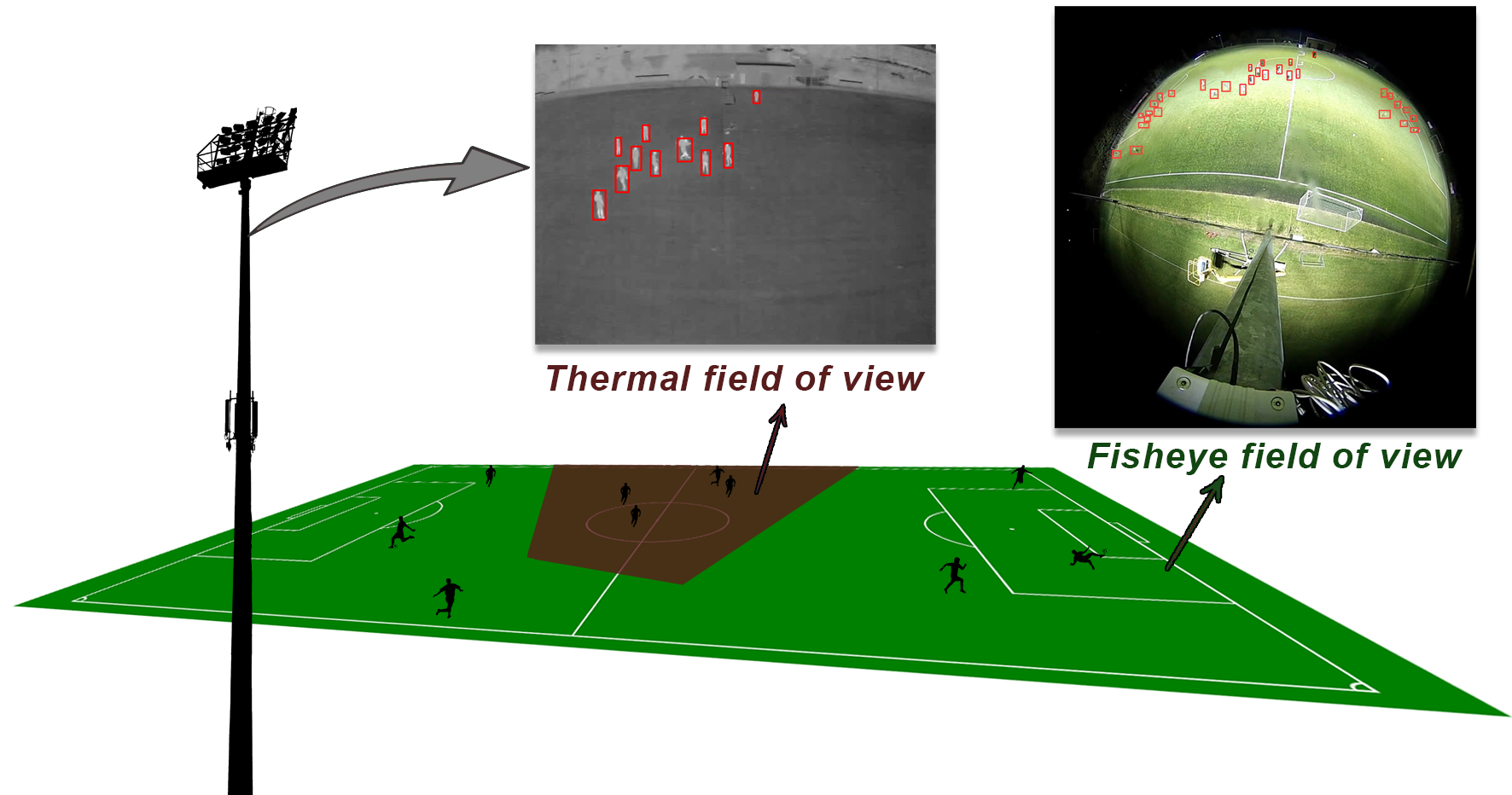}
    \caption{Illustration of the problem handled in this paper. We leverage the detections made on a thermal image on a part of the field to detect all the players on the whole field on the fisheye image.}
    \label{fig:ImEx}
\end{figure}

\section{Introduction}

Local sports fields can be expensive to construct and maintain, especially those built with artificial turf. Therefore, it is important to monitor and then optimize the occupancy of existing fields and stadiums. Furthermore, an automatic occupancy analysis method may open up new possibilities within real-time information and booking. In this work we propose a robust and cost-effective method for player detection and counting in a football field. 

For robust video monitoring of outdoor football fields, one main challenge is the size of the field. A field may be covered by either several regular cameras, which makes the setup rather complex and expensive, or it is possible to use a camera with a wide field of view, such as a fisheye camera. However, with a fisheye camera covering the entire football field, the players will appear small and have different orientation in the image due to the lens distortion. Player detection on these types of images is therefore not a trivial task. 
Another main challenge in outdoor environments is varying lighting conditions. Even though a football field may be illuminated during nights, lighting conditions will change during the day due to changing weather, position of the sun, and the effect of artificial lighting.
To avoid problems with difficult lighting conditions, thermal cameras may be considered. 
These cameras capture only thermal infrared radiation, which represents temperature in the scene, hence they are more independent of lighting and normally eases the task of person detection because people have a temperature different from the background \cite{Gade2014}. However, thermal cameras are expensive and due to their limited field of view and resolution, several cameras would be needed to cover a football field.

To construct a camera setup that is reasonable in price level and at the same time robust to changes in weather and lighting conditions, we propose to use one fisheye RGB and one thermal camera co-located at the side of the field. An illustration of the setup and example images from the two cameras are shown in Figure~\ref{fig:ImEx}. Only the fisheye camera will cover the entire field, while the detections obtained directly from the thermal camera will serve to provide some kind of ground truth for teaching a network.


There are two main contributions in this paper: (i) We show how two different image modalities and fields of view can be combined in a student-teacher distillation approach. (ii) We show how a student network can be trained to detect players outside the field of view of the teacher, through a combination of a custom data augmentation process and a motion detection algorithm. 
\section{Related work}
\mysection{Player detection in sports.} Detection of players in sports fields is the first step of vision systems for sports applications, like occupancy analysis, tracking, performance analysis, etc. \cite{THOMAS20173}. Background subtraction based methods have often been used for player detection due to the fast processing time that makes it well-suited for real-time applications. It has been applied for static cameras \cite{Archana16, Reno15} and for moving cameras in the case of uniformly colored surfaces \cite{Rao15}. However, noise should be expected due to, \eg, other moving objects, similar colors in foreground and background, changing lighting conditions, and shadows. It has also been proposed to use classic person detection methods like using the AdaBoost algorithm for training a linear classifier with HOG features for detecting players in Australian Rules Football \cite{Faulkner15}, or similarly with AdaBoost and Haar features for player detection in basketball \cite{Ivankovic12} and baseball \cite{Mahmood12}.

More recently, like for general object detection, CNN-based methods have also been the dominant trend for detecting sports players. In \cite{Sah19} a shallow CNN was trained to detect players on a hockey field, while others use pre-trained networks like Mask R-CNN for handball videos \cite{Pobar18} and basketball videos \cite{Yang18}, or YOLO for handball videos \cite{Buric19}. In \cite{Zhang18} a reverse  connected  convolutional  neural  network  (RC-CNN) is proposed for player detection. The reverse connected modules are embedded into the CNN to pass semantic information captured by deep layers back to shallower layers.

\mysection{Person detection in fisheye and thermal cameras.} Fisheye cameras have been widely used for person detection because of their advantage of wide viewing angle. Methods using a single camera setup have been reported for surveillance \cite{kim1, kim}, automobiles \cite{levi}, indoor environment \cite{sait, wang} and outdoor sports field \cite{hudafish}. In these methods, the setup was used for pedestrian detection, tracking and occupancy analysis. Multiple camera setups are also proposed to detect persons for similar applications \cite{bert, ngu, wang1}. However, the main disadvantages with fisheye cameras are the distortion on the borders and the lower image quality in low lighting conditions.

Thermal cameras have long been used in practice because of their efficiency in bad lighting conditions. The range of applications varies from industrial uses to daily life traffic and surveillance \cite{Gade2014}. Various methods based on thermal cameras have been proposed for person detection, such as feature extraction and threshold based methods  \cite{Dai, GadeO, GadeLO, Zhang}, HOG methods \cite{Li2, tumas}, machine learning techniques \cite{Huda} and deep neural networks \cite{heo, Herrmann,Noor2020TheEffect}. A dataset and a trained network for people detection on outdoor thermal images have been proposed in \cite{Noor2020TheEffect}. The disadvantage of thermal cameras is their expensive cost and their reduced field of view.

In this work we will continue on recent trends to use a CNN-based method for player detection. We aim to circumvent the limitations of both fisheye and thermal cameras, by combining these modalities and teach the network for the fisheye camera with detections from the thermal camera, in a student-teacher distillation approach.

\section{Data acquisition and calibration}

\mysection{Camera setup.} The data used in this work consist of video streams of two different cameras: a fisheye camera and a thermal camera. 
Both cameras are installed on the same pole at the side of a football field, as illustrated in Figure~\ref{fig:ImEx}. 
The thermal camera is placed approximately $9.8$ meters above the ground and the fisheye camera is installed at $9.5$ meters. By doing so, the field of view of the fisheye camera covers the whole football field, whereas the thermal camera covers the central area, as shown in Figure~\ref{fig:ImEx}. 
In this setup, the field of view of the thermal camera represents $6\%$ of the fisheye image, and covers $22\%$ of the football field as seen by the fisheye camera.
Let us note that several teams use the field simultaneously for a training session during the video. Hence, the players are performing different activities, such as moving goals or performing various exercises. Therefore, the players can be found in different postures in any part of the field.

\mysection{Acquisition.} The fisheye video stream is recorded using a Hikvision Fisheye Network Camera with a resolution of $1280\times1280$ pixels and a field of view of $360^\circ$. The thermal video stream is recorded using an Axis Q1922 camera that has a resolution of $640\times480$ pixels and $57^\circ$ of horizontal viewing angle. 
The videos were recorded during one hour in an amateur football field in December 2017, at night time with artificial light illuminating the field. The fisheye camera records the video at $12$ fps. The thermal camera initially records the video at $30$ fps, which is then re-sampled at $12$ fps to allow a synchronization of the two streams. 
A proper camera calibration and registration between fisheye and thermal images is required for the transferability of points of interest. 



\mysection{Calibration and registration.} First, a calibration of the internal parameters of each camera is performed following the procedure described in \cite{chris}. For the thermal camera, an A3-sized 10 mm polystyrene foam board is used as backdrop and a board of the same size with cut-out squares is used as checkerboard. In order to obtain a suitable contrast, the backdrop is heated and the checkerboard is placed at room temperature before the calibration. For the fisheye camera calibration, a checkerboard of $25\times25$ centimeters is used. Finally, the camera parameters derived from the calibration are obtained with a Matlab toolbox \cite{camera}.



\begin{figure}
\centering
\includegraphics[width=0.95\columnwidth]{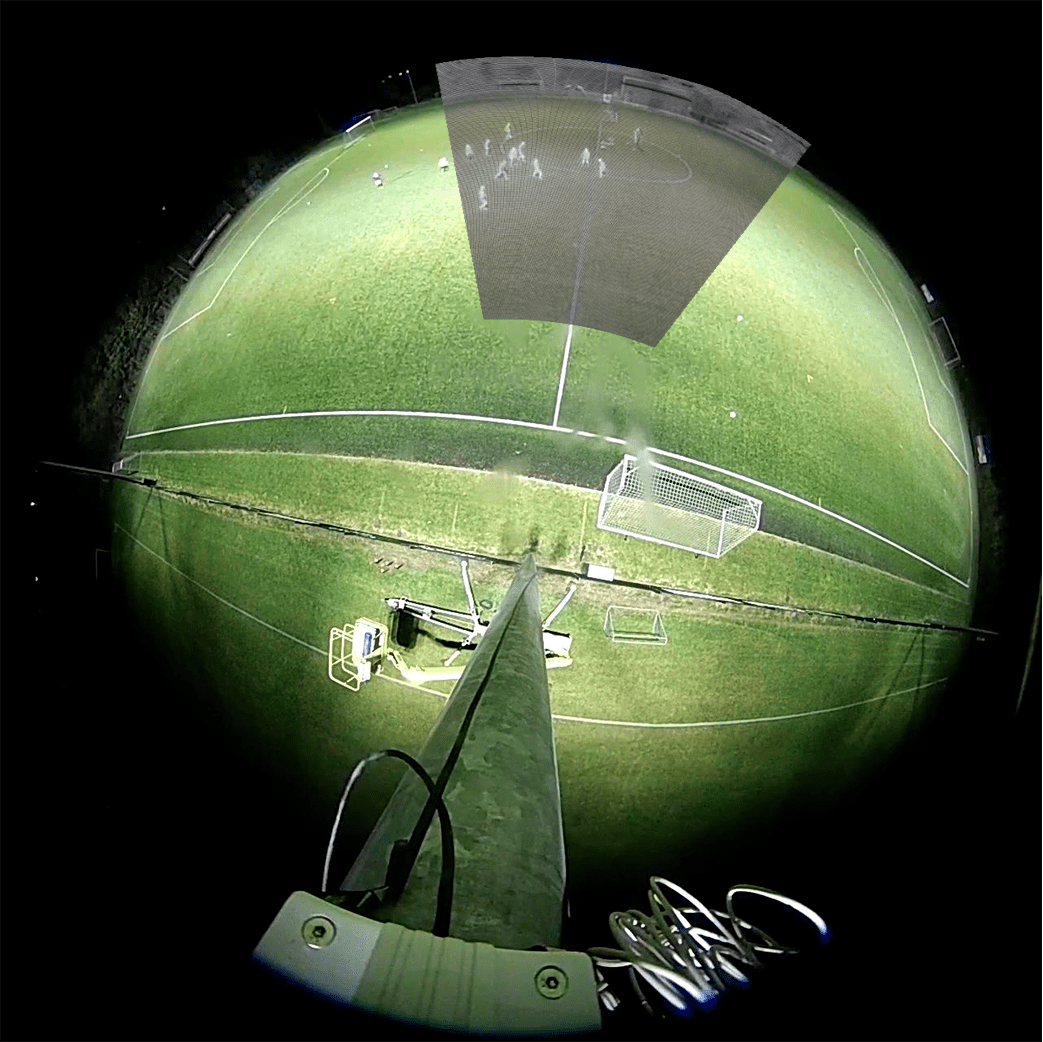}
\caption{Projection of the thermal image onto the fisheye image. The thermal camera sees only $\approx 22\%$ of the football field pixels of the fisheye image.}
\label{fig:mapped}
\end{figure}

Second, we perform the registration between the two cameras. We undistord the images of the cameras using the internal parameters obtained previously. We manually choose several points of interest on the undistorded football field to compute the homography between the cameras, following \cite{registration}. These points are player feet positions for the players seen by the two cameras. 
The projection of the thermal image onto the fisheye image is shown in Figure~\ref{fig:mapped}.

\section{Methodology}

\mysection{Problem statement.} A general formulation of the problem tackled in this paper is the following. Given a network performing a detection task on data from a camera, how can we train a real-time network for the same detection task on data from another camera with a possibly different modality and a different field of view of the same scene? 
In this section, we describe our solution for this problem in general terms, and we also explain how each step is particularized for our practical use case. Our use case consists in the task of player detection on a football field given a network able to detect players on a fixed thermal camera with a narrow field of view, which is used to train another detection network on data from a fixed fisheye camera with a wide field of view. This is illustrated in Figure~\ref{fig:ImEx}.

\mysection{Notations.} We handle this problem with a teacher-student distillation approach, in which the output of a trained teacher network \teacher serves as surrogate ground truth to train a student network \student (see \cite{Wang2020Knowledge} for a recent review). Such a method has already been successfully applied in sports in \cite{Cioppa2019ARTHuS} for segmenting football and basketball players in real time by distilling a slow \teacher (Mask R-CNN \cite{He2017MaskRCNN}) into a fast \student (TinyNet \cite{Cioppa2018ABottomUp}). In addition, in \cite{Cioppa2019ARTHuS}, the distillation is performed in an online fashion, such that \student continuously adapts to the latest game conditions. However, \teacher and \student use the same video feed, which implies that \student can be directly (no transformation needed) and entirely (no missing ground truth) supervised by \teacher. 

\begin{figure}
\centering
    \includegraphics[width=0.9\linewidth]{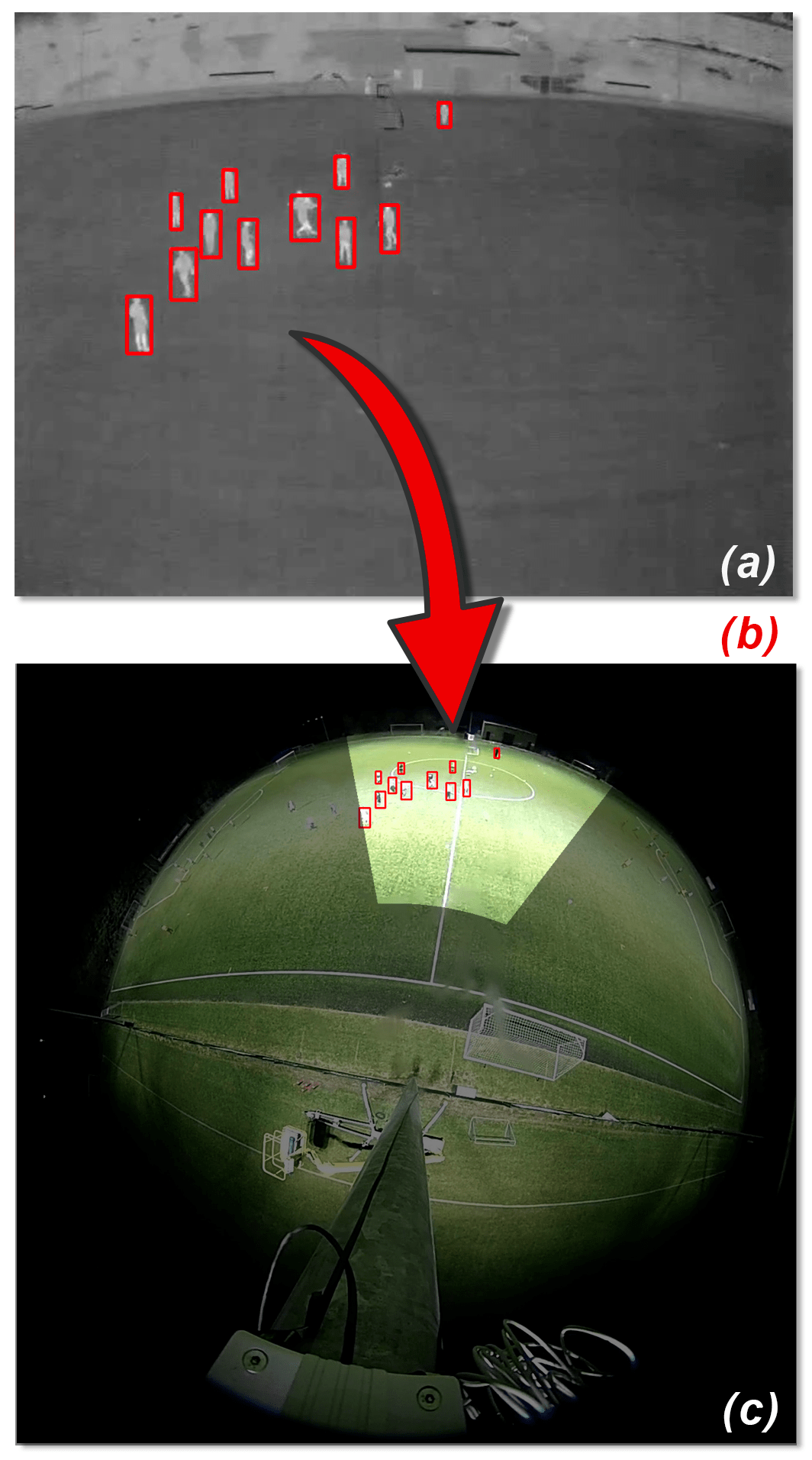}
    \caption{The bounding boxes given by \teacher on \frameteacher (a) are projected (b) into \framestudent to provide us surrogate ground-truth bounding boxes in \intersectFOV (c).}
    \label{fig:GT-in-P}
\end{figure}

In the present work, the setup is more challenging as \teacher and \student process the video feeds of two cameras \camteacher and \camstudent with different modalities and fields of view. Having different modalities prevents us from using \teacher on the feed of \camstudent, and having different fields of view prevents us from directly and entirely supervising \student. We assume that \camteacher and \camstudent are synchronized, such that they capture frames \frameteacher and \framestudent simultaneously at each capture time \capturetime. We also assume that the projection from \frameteacher to \framestudent, expressed in terms of pixel coordinates, is known from the preliminary calibration step explained in the previous section. We note \intersectFOV the area of \framestudent representing the projection on \framestudent of the part of the scene also filmed by \camteacher (shown in Figure~\ref{fig:GT-in-P}). The remaining part of \framestudent is filmed by \camstudent only and is noted \notintersectFOV. As both cameras are fixed, this partition of \framestudent is independent of \capturetime. 

In order to train \student, we need surrogate ground-truth bounding boxes both in \intersectFOV and in \notintersectFOV.  We detail hereafter how we obtain such boxes in \framestudent for a given capture time \capturetime. Following common practice, we represent a bounding box coordinates by a quadruplet containing the two coordinates of the center of the box, its width and its height.

\mysection{Surrogate ground truths in \intersectFOV.} This part is straightforward. First, we use \teacher to detect players in \frameteacher and retrieve the coordinates 
of bounding boxes of \frameteacher. Then, we project them into \framestudent using the calibration of the previous section. By doing so, we obtain the surrogate ground-truth bounding boxes of \framestudent that are located in \intersectFOV, as shown in Figure~\ref{fig:GT-in-P}. The remaining part of \intersectFOV constitutes detection-free areas.

\mysection{Surrogate ground truths in \notintersectFOV.} This part is more difficult as we cannot have a direct access to the pixels of \notintersectFOV from those of \frameteacher. Training \student solely with the boxes provided in \intersectFOV for each \framestudent leads the network to focus only on \intersectFOV and to overlook \notintersectFOV for each frame. Eventually, the network is not able to detect anything in \notintersectFOV. 

To circumvent this problem, our idea is the following. First, we use a custom data augmentation process to create artificial players with known bounding boxes in \notintersectFOV. This provides us the ``ground-truth locations'' of some ``true positive'' players that \student will have to detect. This is not sufficient as we still need ``ground-truth information'' in areas where we did not create any player. For that purpose, we use a motion detection algorithm to identify areas of \notintersectFOV that are guaranteed player-free. This provides us ``true negative'' areas, in which \student will be penalized when predicting player bounding boxes. In the remaining areas of \notintersectFOV, we have no useful information, hence \student will not be penalized. These two steps are described in detail hereafter.

\begin{figure*}
    \centering
    \includegraphics[width=0.98\textwidth]{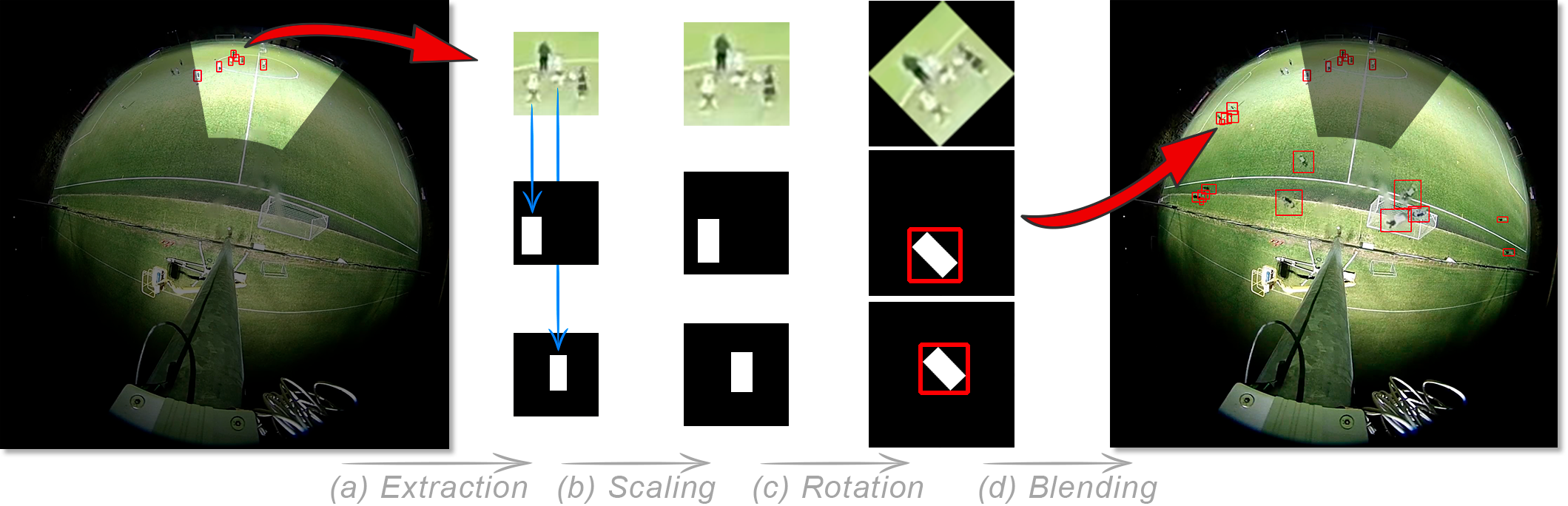}
    \caption{Our custom data augmentation pipeline designed to construct surrogate ground-truth bounding boxes in the region \notintersectFOV filmed by \camstudent only. First, crops containing players are extracted (a) from the area filmed by both cameras \intersectFOV, in which we know their location. Then, each crop and its associated bounding boxes are scaled (b) and rotated (c) to be appropriately pasted in \notintersectFOV. A seamless blending is applied during the collage to increase the realistic aspect of the augmented image. As a result, we create artificial players with known bounding boxes in \notintersectFOV.}
    \label{fig:data-augmentation}
\end{figure*}

\mysection{[1. Custom data augmentation]} In order to introduce true positive players with known bounding boxes in \notintersectFOV, we design the following automatic data augmentation process. Given a frame \framestudent, we start by randomly extracting image crops delimited either by one isolated or by several adjacent bounding boxes previously obtained in \intersectFOV (Figure~\ref{fig:data-augmentation}). Then, for each crop, we randomly select a pixel in \notintersectFOV, which will serve as an anchor point where the crop will be pasted after being rescaled and rotated appropriately. In our use case, the anchors are selected in the subset of \notintersectFOV corresponding to the football field.

We perform a rescaling and a rotation of each crop to produce an insertion that looks as realistic as possible by taking into account the inherent distortions of \camstudent (Figure~\ref{fig:data-augmentation}). Let \polarinitial denote the initial polar coordinates (with origin located at the center of \framestudent) of the center of the crop and \polarfinal those of its selected anchor point. We rescale the crop by a factor $\alpha\me^{\beta(\radiusfinal-\radiusinitial)}+\gamma$ with $\alpha =0.5 ,\beta=-0.004,\gamma=0.5$ and rotate it by the angle difference $\anglefinal-\angleinitial$. Finally, we paste the transformed crop on \framestudent itself with OpenCV's seamless blending function, such that its center is located at the selected anchor point (Figure~\ref{fig:data-augmentation}). In order to obtain the boxes associated with these artificial players, we perform the same transformation on each bounding box included in the initial crop. Eventually, for each transformed box, we consider as surrogate ground-truth bounding box the smallest unrotated (regular) rectangular box that encloses it (Figure~\ref{fig:data-augmentation}). 

In our fisheye setup, the data augmentation process allows to create artificial players with known bounding boxes in \notintersectFOV (Figure~\ref{fig:data-augmentation}). However, this does not suffice to train \student efficiently, as real players without known boxes may still be present in \notintersectFOV. In a standard training process, \student would thus be forced to detect the artificial players and would be penalized for detecting the remaining real ones. To bypass this undesirable effect, we remove the penalty suffered by \student for detections containing enough motion. Hence, we leverage a motion detection algorithm to determine where this should be applied. By doing so, we also obtain areas where there is assuredly no player, \ie where detections should not be made.

\begin{figure}
\centering
\includegraphics[width=0.99\columnwidth]{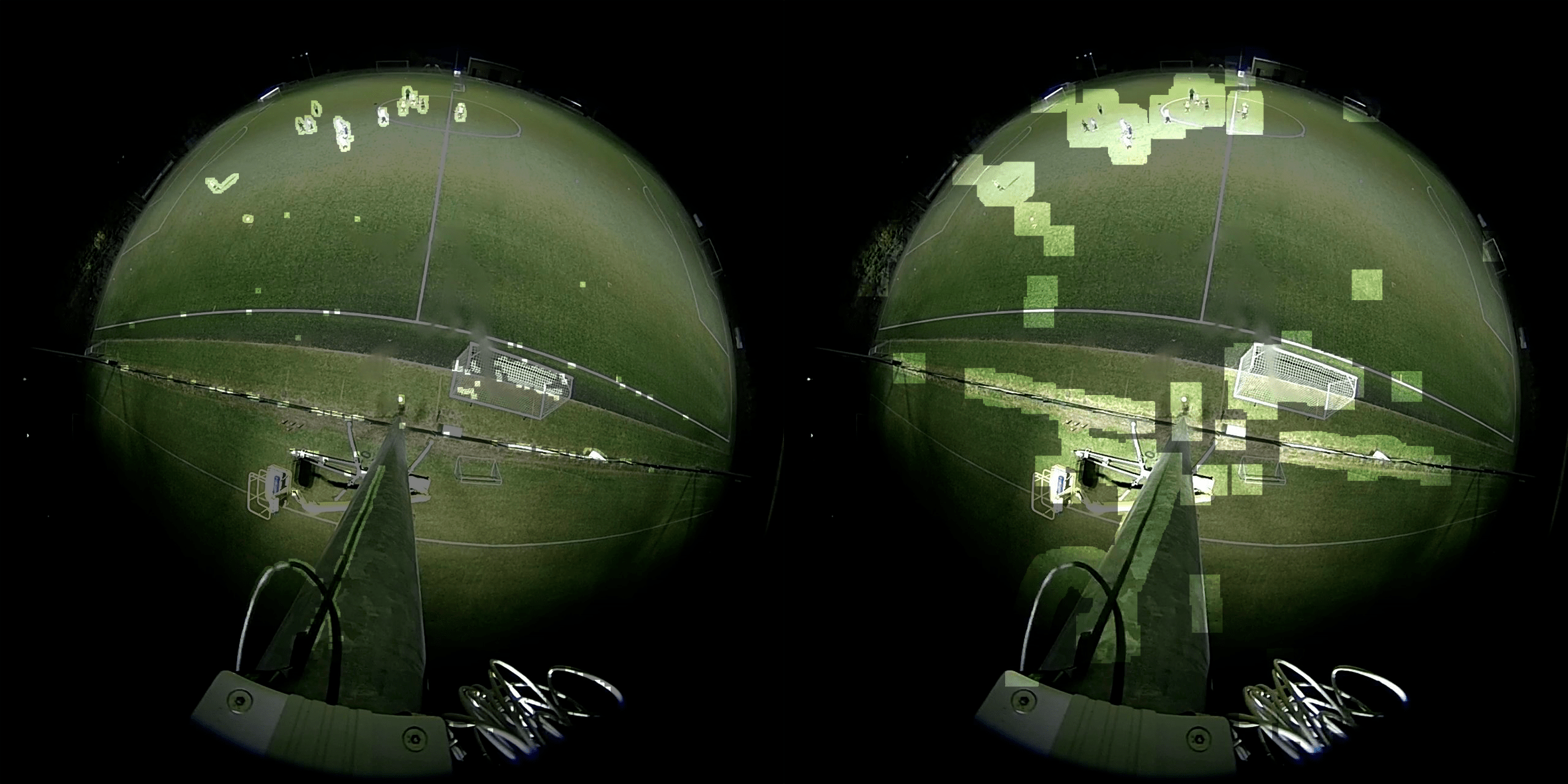}
\caption{Initial motion detection mask \maskVibe overlayed on its corresponding frame (left), and enlarged motion detection mask \bigmaskVibe (right).}
\label{fig:vibe-mask}
\end{figure}

\mysection{[2. Motion detection]} As we handle a video feed from a fixed camera, we use ViBe \cite{Barnich2011ViBe} to obtain, for each frame \framestudent, the set of pixels that are in motion, noted \maskVibe, and those that are not, noted \notmaskVibe (Figure~\ref{fig:vibe-mask}). 
ViBe is very sensitive to motion, which implies that, in our fisheye setup, \maskVibe almost surely contains all the players, as well as pixels corresponding to the balls, player shadows, and some noise. As \maskVibe may be tight around the players, we morphologically dilate it by a $11\times 11$ square kernel to ensure that it includes the bounding boxes that would surround the players if they were available (Figure~\ref{fig:vibe-mask}). By doing so, we obtain an enlarged mask \bigmaskVibe, such that we can penalize \student when it detects players in \notbigmaskVibe, \ie outside the enlarged mask. However, \bigmaskVibe remains an area of uncertainty, where we do not penalize \student. Technically, this means that we zero out the loss in this area during training, as detailed hereafter.


\mysection{Training \student.} We use the YOLOv3 network~\cite{Redmon2018YOLOv3} trained to detect humans on thermal images in~\cite{Noor2020TheEffect} as teacher network \teacher. We use YOLOv3-tiny \cite{Redmon2018YOLOv3} as student network \student, adapted for a single class problem and with four times less channels for each convolutional layer. Hence, \student outputs a list of $5$-dimensional vectors. Each of them encapsulates information on a predicted bounding box: the four coordinates \coordstudent defining the box, and a player score \objectness representing its confidence for a player to actually belong to the box. 

\begin{figure}
\centering
\includegraphics[width=0.95\columnwidth]{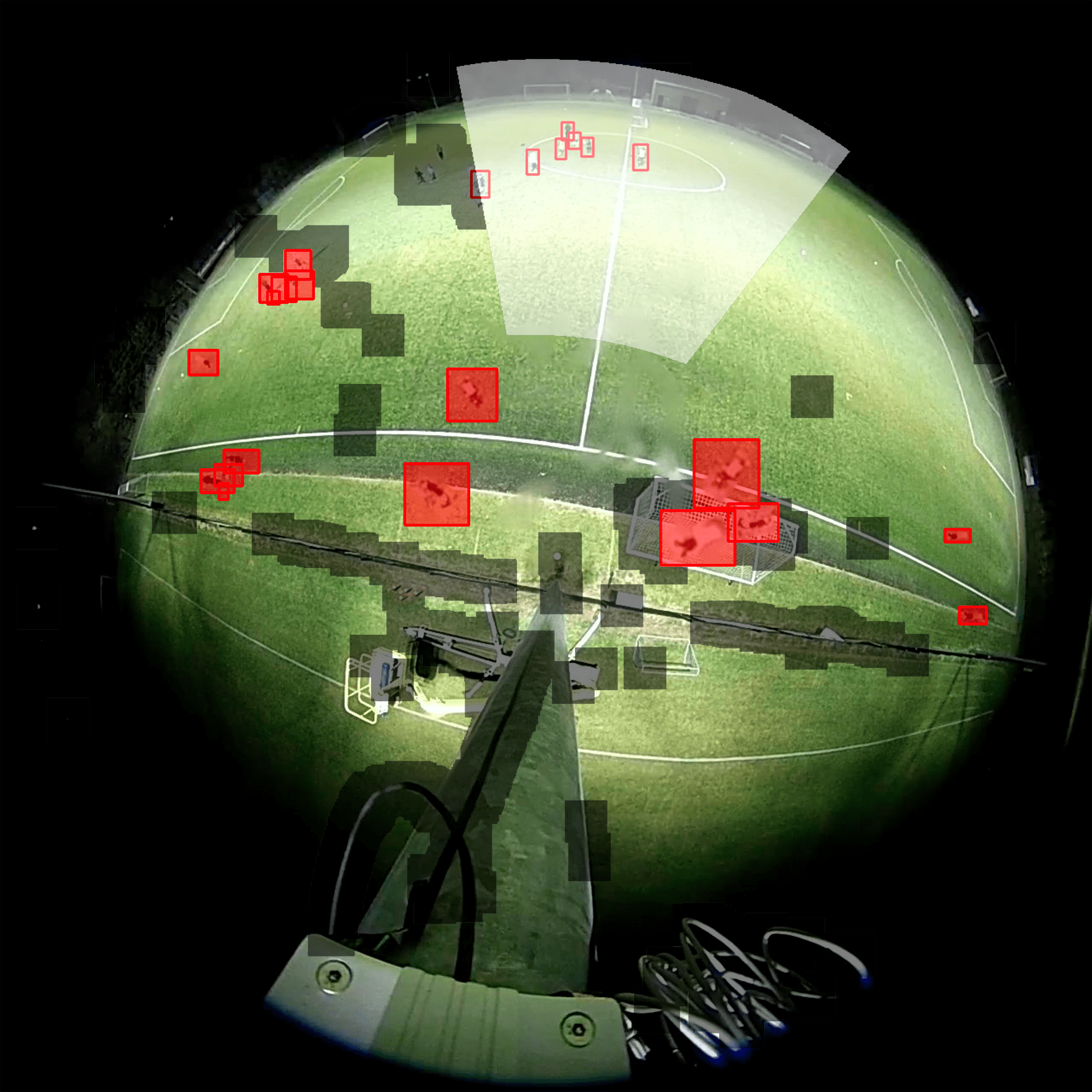}
\caption{Combination of our data augmentation and motion detection algorithms, showing how the loss is applied to penalize the predictions of \student in \notintersectFOV (outside the white area). \student must detect the players artificially created (red rectangles). Also, predicted boxes whose center falls within the enlarged motion mask \bigmaskVibe (the black zones) do not generate any loss, since this area includes the players of \notintersectFOV not erased by the data augmentation, for which we have no ground-truth boxes. Finally, \student must not predict any box in the rest of the image in \notintersectFOV. Let us recall that the loss is applied everywhere in \intersectFOV, as we have the ground truth from \teacher in that area.}
\label{fig:no-loss}
\end{figure}

The loss of YOLOv3-tiny, hence \student, penalizes these vectors in the following way (see Figure~\ref{fig:no-loss}). For a predicted box close to a surrogate ground-truth box (either in \intersectFOV or in \notintersectFOV), the mean square error loss between the coordinates of the boxes is computed, as well as the binary cross-entropy loss of \objectness. This encourages the network to predict a high confidence score (closer to $1$) and to find the right dimensions of the box. For a box far from a surrogate ground-truth box, only the binary cross-entropy loss of $1-\objectness$ is computed, to discourage the network from predicting a player in that box (\objectness closer to $0$). In our case, we must take into account the uncertainty about the boxes in \bigmaskVibe in the region \notintersectFOV, as they may correspond to unnanotated real players. Therefore, for a box far from a surrogate ground-truth box (including those created by the data augmentation), we zero out its loss if the center of the box is in \notintersectFOV and is in motion (belongs to \bigmaskVibe). If the center of the box belongs to \notbigmaskVibe, we are practically sure that there is no player in the box, and we thus leave the loss as is to penalize that detection. There is not particular restriction about the loss in \intersectFOV. This is illustrated in Figure~\ref{fig:no-loss}.

\mysection{Inference.} When used for inference, we verify that the bounding boxes predicted by \student contain enough motion. Indeed, the predicted boxes whose center is not in motion, \ie outside \bigmaskVibe, are not likely to contain a player. Therefore they are removed from the final output of \student.
\section{Experiments}

\mysection{Online distillation.} 
In this work, we perform the distillation of the teacher network \teacher into the student network \student in an online manner as in \cite{Cioppa2019ARTHuS}. The reason for using that process is threefold. First, this allows \student to continuously adapt to the latest weather and lighting conditions. Second, in a real-life deployment of the system, the online distillation will indeed be performed continuously. Hence, in order to have an understanding of how \student behaves as it trains and detects people in real time, it is worth testing \student under similar conditions. Third, training \student adaptively allows us to study the evolution of the performance of the network as it learns through time. As we have only one video sequence with both the thermal and the fisheye recordings, this also enables us to evaluate \student multiple times rather than measuring its performance only once, on a unique (and maybe abnormally hard or easy) small set of frames. 

In the online distillation process, all the frames of the fisheye camera \camstudent are treated by \student, which runs in real time. Meanwhile, some frames of the video feed of the thermal camera \camteacher are input to \teacher, which provides boxes converted into surrogate ground-truth bounding boxes in the area \intersectFOV of the frame captured by \camstudent. These boxes are accumulated in an online dataset with $5$-minutes memory, and the dataset is used to train a copy of \student in a separate thread. The training is performed on the whole frames \framestudent as described in the previous section, using our data augmentation and motion detection processes outside \intersectFOV. When this copy of \student has trained during one epoch on the online dataset, its weights are updated and transferred into the initial network \student that performs the detection on all the frames. Consequently, the weights of this network evolves through time to continuously adapt to the latest video conditions.

\mysection{Quantitative evaluation.} To assess the performance of the student network \student over the course of the video, we manually annotated the ground-truth bounding boxes for all the players of one frame every $10$ seconds of the fisheye video. We compute the performance of \student on a set of frames with the Average Precision (AP) metric 
particularized for one class. Following practice for the Pascal VOC dataset \cite{Everingham2010PascalVOC}, each bounding box predicted by \student is matched with the ground-truth box with which it has the largest intersection over union (IoU). We consider predicted boxes with an IoU larger than some threshold \tiou as true positives, the others as false positives, and the ground-truth boxes left unmatched are false negatives. If several true positives are associated with the same ground-truth box, only one of them is kept as a true positive, while the others are rather considered as false positives. We note the number of true positives (resp. false positives, false negatives) TP (resp. FP, FN). Then, we compute the precision and recall as
\begin{equation*}
    P=\frac{TP}{TP+FP}\hspace{0.5cm} \text{and} \hspace{0.5cm} R=\frac{TP}{TP+FN}.
\end{equation*}
We compute the points $(P,R)$ for various thresholds on the confidence scores of the boxes to obtain the PR curve. Finally, we compute the area under the PR curve as suggested in \cite{Everingham2010PascalVOC} to obtain the AP for that set of frames. Despite its limitations \cite{Boyd2012Unachievable}, this kind of evaluation process has been widely adopted in the community.

\begin{figure}
\centering
\includegraphics[width=0.95\columnwidth]{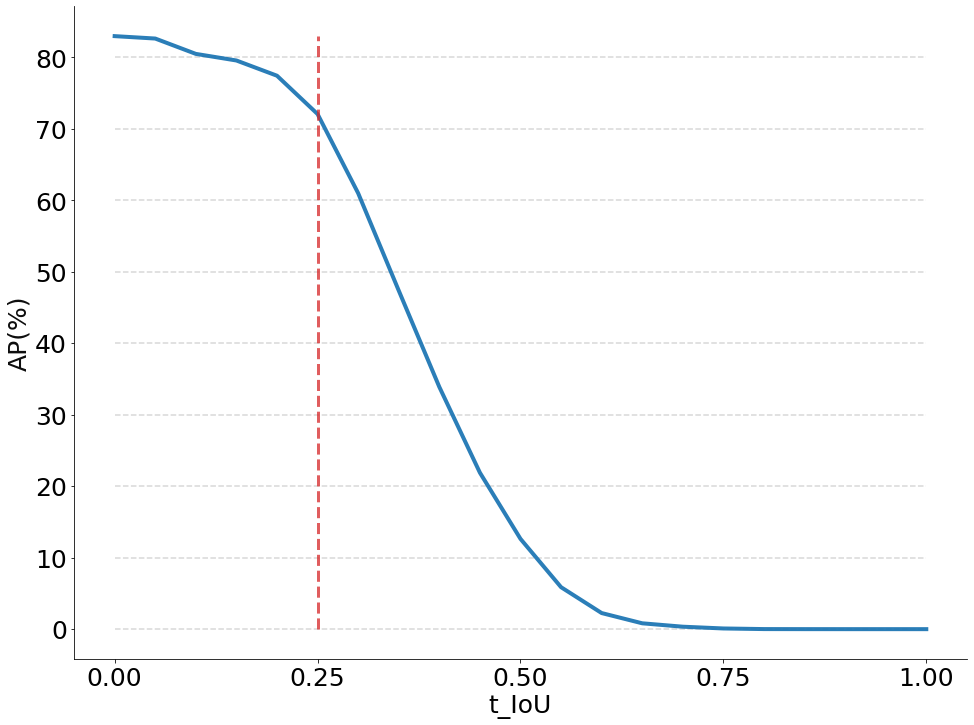}
\caption{\textcolor{anthoblue}{\textbf{Performances of \teacher in \intersectFOV}} on the last $15$ minutes of video as a function of \tiou. This quantifies how accurately \teacher centers its bounding boxes on the players. We can see that \teacher is not perfect. We decide to evaluate the \textcolor{anthored}{\textbf{performances of \student for \tiou$=0.25$}}, as we consider it as the largest \tiou for which \teacher still displays satisfying performances (AP $>70\%$).}
\label{fig:teacher-perf}
\end{figure}

In order to determine an appropriate value of \tiou for evaluating the performance of \student, we examine the efficiency of \teacher in predicting the boxes in \intersectFOV. For that purpose, we compute the AP of \teacher on the last $15$ minutes of video, for several values of \tiou ranging from $0$ to $1$, for the frames where ground-truth annotations are available. This allows us to determine how good \teacher is at centering its bounding boxes on the players. The performance of \teacher in \intersectFOV as a function of \tiou is shown in Figure~\ref{fig:teacher-perf}. We can see that \teacher is not perfect in \intersectFOV, which conditions the performances that can be expected from \student. To evaluate \student, we choose \tiou$=0.25$, as \teacher displays reasonable performances in \intersectFOV with that threshold. Given the small size of the boxes, it also makes sense to examine the performance of \student for a relatively low value of \tiou. Let us recall that the boxes outputted by the network are independent of any particular choice of threshold. It serves only for quantitative evaluation purposes.

\begin{figure}
\centering
\includegraphics[width=0.95\columnwidth]{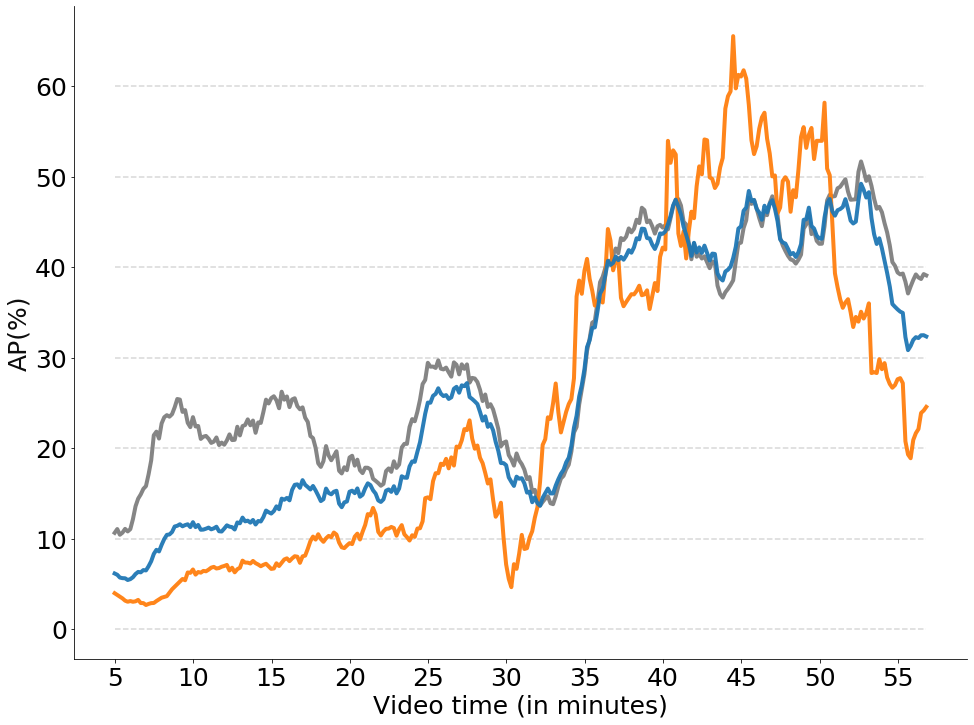}
\caption{Evolution of the performances of the student network \student through the video in \textcolor{anthogray}{$\pmb{\intersectFOV}$}, \textcolor{anthoorange}{$\pmb{\notintersectFOV}$}, and in the \textcolor{anthoblue}{\textbf{whole frames}}. We can see that the network improves over time and that it manages to perform well both in \intersectFOV and in \notintersectFOV.}
\label{fig:perfs}
\end{figure}

Following \cite{Cioppa2019ARTHuS}, we evaluate the performance of the student network \student progressively. 
Every $10$ seconds, \student predicts the bounding boxes of the frames for which we have manual annotations within a running temporal window that covers the next $3$ minutes of video. For this set of frames, we compute the AP. The evolution on the AP through time with \tiou$=0.25$ is represented in Figure~\ref{fig:perfs}. We see that the performance tends to increase, indicating that \student learns to better detect players over time. Figure~\ref{fig:perfs} also reveals that there is still room for improvement in the present challenge.

We further examine the effectiveness of our data augmentation and motion detection processes to train \student for detecting players outside \intersectFOV. For that purpose, we perform a region-specific analysis by computing the temporal evaluation of the AP within \intersectFOV and \notintersectFOV. The performance curves are displayed in Figure~\ref{fig:perfs}. We note that \student learns efficiently to detect players in \notintersectFOV, as the performances for \intersectFOV and \notintersectFOV are close to each other and follow the same trend. Also, further experiments reveal that the post-processing with the motion mask \bigmaskVibe is particularly helpful to increase the performance in \notintersectFOV. In that area, the AP decreases by $5$ to $20\%$ without post-processing, while the drop is below $3\%$ in \intersectFOV.

\begin{figure}
\centering
\includegraphics[width=0.95\columnwidth]{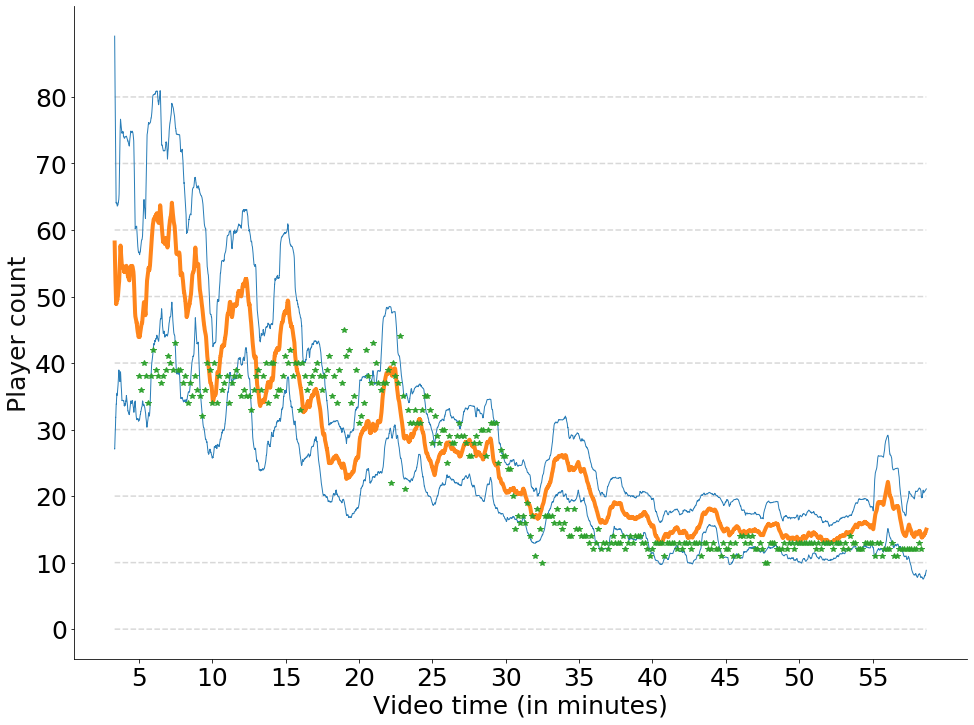}
\caption{\textcolor{anthoorange}{\textbf{Results on the player counting task}} averaged over a $1$-minute window, and associated \textcolor{anthoblue}{\textbf{standard deviation}}. During the last $15$ minutes, we have a RMSE with the \textcolor{anthogreen}{\textbf{ground truth}} of $3.4$ players, which is reasonable and shows that our method provides a reliable estimate of the occupancy of the football field.}
\label{fig:counting}
\end{figure}

Finally, as a potential application of this system is to monitor the use of the football field, we examine the results obtained for the task of people counting. The predicted number of people on the field corresponds to the number of bounding boxes predicted by \student (thus on the fisheye images) after post-processing. We average the counting using a $1$-minute sliding window. The results are displayed in Figure~\ref{fig:counting}. We note that our method gives a globally reliable estimate of the number of people present on the field. Quantitatively, during the last $15$ minutes of video, the root mean square error (RMSE) between the predictions and the ground truth is as low as $3.4$ players. Again, we can see that the performance tends to increase over time since the estimate is more accurate at the end of the video, indicating that \student learns to better detect players over time. Also, we can see in Figure~\ref{fig:counting} that the standard deviation of the box count computed for each sliding window decreases over time, which indicates that the network becomes more consistent as it trains. Even though \student tends to slightly overestimate the actual number of players, we can see that it manages to provides a good overview of the use of the field.

\mysection{Qualitative evaluation.} To further assess the usefulness of our data augmentation and motion detection processes, we perform ablation studies on the components of our method. We investigate the combination of either enabling or disabling the data augmentation, with either zeroing out the loss in the motion mask \bigmaskVibe, or nowhere in \notintersectFOV, or everywhere in \notintersectFOV. The effects observed for these setups are reported in Table~\ref{tab:ablation-results}. In our experiments, we observe that the combination of the data augmentation and of zeroing out the loss in \bigmaskVibe, as detailed in this paper, leads to the best student network \student at inference time. Activating the loss everywhere in \notintersectFOV at training time forces \student to detect only the artificial players in \notintersectFOV and to avoid detecting the actual players of \notintersectFOV that have not been erased by the data augmentation. This may confuse \student, leading to a decrease in its ability to detect players in \notintersectFOV at inference time. We notice that canceling the loss everywhere in \notintersectFOV leads to thousands of predicted bounding boxes in \notintersectFOV at inference time. This makes sense since the network is not forced to detect or not players in \notintersectFOV in this case. Most of these predictions are false positives, and the system is useless in practice. As indicated in Table~\ref{tab:ablation-results}, we also note that removing the data augmentation always leads to mediocre networks, for similar reasons as those already explained. In particular, activating the loss everywhere in \notintersectFOV makes \student unable to detect any single player in \notintersectFOV. This results from the absence of ground-truth true positives (both artificial and real ones) in \notintersectFOV.

\begin{table}[t]
    \centering
    \resizebox{\columnwidth}{!}{%
    \begin{tabular}{c||c|c}
         In \notintersectFOV & \makecell{With data \\ augmentation} & \makecell{Without data \\ augmentation} \\ 
         \hline
         \hline
        \makecell{Cancel loss in \\ the motion \\ mask \bigmaskVibe} & \makecell{\textbf{Our full method.} \\ Most players in \notintersectFOV \\ correctly detected, \\ few false positives.} & \makecell{Few players \\ detected in \notintersectFOV, \\ unusable in practice} \\ 
        \hline
        \makecell{Activate loss \\ everywhere \\ in \notintersectFOV} & \makecell{Able to detect \\ players in \notintersectFOV, \\ but not as good as \\ our full method} & \makecell{Unable to make\\ any detection in \notintersectFOV,\\ no true positives} \\ 
        \hline
        \makecell{Cancel loss \\ everywhere \\ in \notintersectFOV} & \makecell{Thousands of \\ detections in \notintersectFOV, \\ mostly false positives} & \makecell{Thousands of \\ detections in \notintersectFOV, \\ mostly false positives} \\
    \end{tabular}
    }
    \caption{Ablation results in \notintersectFOV. The combination of the data augmentation and the motion detection algorithm gives the best trade-off between true and false positive detections.}
    \label{tab:ablation-results}
\end{table}

\begin{figure}
\centering
\includegraphics[width=0.95\columnwidth]{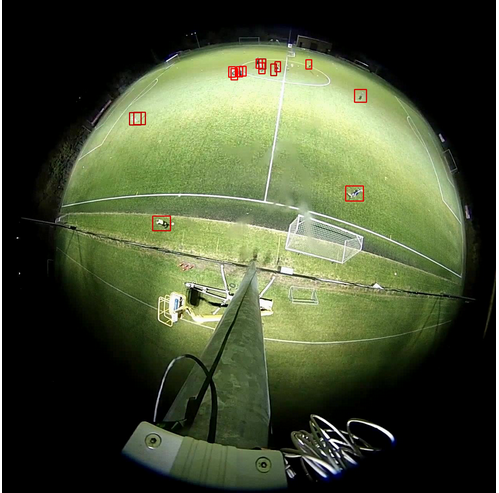}
\caption{Detections on a test frame. We can note that players are accurately detected, even though there are a few superfluous predicted bounding boxes.}
\label{fig:example-result}
\end{figure}

Finally, examples of detections provided by \student are given in Figure~\ref{fig:example-result}. We can see that players located in \notintersectFOV are detected as efficiently as those located in \intersectFOV. This was made possible thanks to our data augmentation and motion detection algorithms in the distillation approach.

\section{Conclusion}

In this work, we propose a novel system for monitoring the field occupancy in low-budget football stadiums. Our system uses a single wide-angle fisheye camera assisted by a thermal camera to detect and count all the players on the field. We use a network trained in a student-teacher distillation approach. The student network is locally supervised by a teacher network that easily detects players on the thermal camera. These detections are then projected into the fisheye camera using camera registration and serve as surrogate ground truths. Since both cameras have different modalities and fields of view of the scene, the student cannot be fully supervised by the teacher. Therefore, we develop a custom data augmentation process, combined with motion information provided by a background subtraction algorithm, to introduce surrogate ground truths outside their common field of view. In our case, we perform the distillation in an online fashion, \ie our student is continuously trained to adapt to the latest video conditions, while performing the player detection in real-time. We show that our system is able to accurately detect players both inside and outside the common field of view, thanks to our custom supervision.


\mysection{Acknowledgments} 
A. Cioppa is funded by the FRIA. A. Deliège is supported by the DeepSport project of the Walloon region, Belgium.

{\small
\bibliographystyle{ieee_fullname}
\bibliography{bibliography}
}

\end{document}